\documentclass[10pt,twocolumn,letterpaper]{article}

\usepackage{wacv}
\usepackage{times}
\usepackage{epsfig}
\usepackage{graphicx}
\usepackage{amsmath}
\usepackage{amssymb}
\usepackage{booktabs}       
\usepackage{multirow}
\usepackage{amsfonts}       

\newcommand{\mb}{\mathbf}

\def\wacvPaperID{****} 

\wacvfinalcopy 

\ifwacvfinal
\def\assignedStartPage{1} 
\fi

\ifwacvfinal
\usepackage[breaklinks=true,bookmarks=false]{hyperref}
\else
\usepackage[pagebackref=true,breaklinks=true,colorlinks,bookmarks=false]{hyperref}
\fi

\ifwacvfinal
\else
\fi

\begin{document}

\title{FRE: A Fast Method For Anomaly Detection And Segmentation}

\author{Ibrahima Ndiour
\and
Nilesh Ahuja 
\and
Utku Genc
\and
Omesh Tickoo \\
}

\maketitle

\begin{abstract}
This paper presents a fast and principled approach for solving the visual anomaly detection and segmentation problem. In this setup, we have access to only anomaly-free training data and want to detect and identify anomalies of an arbitrary nature on test data. We propose the application of linear statistical dimensionality reduction techniques on the intermediate features produced by a pretrained DNN on the training data, in order to capture the low-dimensional subspace truly spanned by said features. We show that the \emph{feature reconstruction error} (FRE), which is the $\ell_2$-norm of the difference between the original feature in the high-dimensional space and the pre-image of its low-dimensional reduced embedding, is extremely effective for anomaly detection. Further, using the same feature reconstruction error concept on intermediate convolutional layers, we derive FRE maps that provide pixel-level spatial localization of the anomalies in the image (i.e. segmentation). Experiments using standard anomaly detection datasets and DNN architectures demonstrate that our method matches or exceeds best-in-class quality performance, but at a fraction of the computational and memory cost required by the state of the art. It can be trained and run very efficiently, even on a traditional CPU. 

\end{abstract}

\section{Introduction}

The goal of anomaly detection is to identify rare, abnormal events such as defects in a part being manufactured from the observation of data. Manual anomaly detection is time- and labor-intensive which limits its applicability on large volumes of data that are typical in industrial settings. Consequently, there is interest in automated solutions to address this problem. There is a long and diverse history of applying machine learning techniques for anomaly detection (see \cite{ruff2021unifying}  for an excellent survey review of anomaly detection research). Recently, there has been a surge of interest in developing deep learning approaches for anomaly detection. These are commonly based on more general out-of-distribution (OOD) detection techniques, which is typically performed by making the network provide an uncertainty or confidence score for each input. An in-distribution input should be predicted with high confidence (low uncertainty) while an OOD input should result in a low-confidence or high-uncertainty outcome. There is a large body of research on providing robust and reliable uncertainty scores from deep neural networks (DNN). This includes the softmax score \cite{hendrycks2016baseline} and its temperature-scaled variants \cite{liang2018enhancing}; Bayesian neural networks \cite{gal2016dropout}; ensembles of discriminative classifiers \cite{lakshminarayanan2017simple}; deep generative models \cite{van2016conditional, hendrycks2019oe, ren2019likelihood}; and parametric class-conditional density modeling over deep features \cite{lee2018simple, ahuja2019bdl_dfm}.

Although, anomaly detection is a type of OOD problem, there are important practical differences. Access to labeled training data is a challenge because anomalies are typically rare occurrences, and hence gathering data that includes a sufficient number of anomalies greatly increases the data collection and annotation effort. Even if it were possible to capture anomalies in the dataset, they would constitute a small fraction of the overall dataset, resulting in a highly imbalanced training dataset. Furthermore, the nature of anomalies can be arbitrary and unknown since failures or defects can occur for a variety of unpredictable reasons, and it may not always be possible to predict a priori the types of anomalies that would be encountered.

Consequently, anomaly detection is commonly posed as a one-class problem in which a distribution of the good, defect-free data is learnt and anomalies are then detected as deviations from this model. Within this framework, various types of models have been proposed in literature. Deep autoencoder based models \cite{zhou2017anomaly} are trained by minimizing the reconstruction error on the normal data and then use the reconstruction error as an indicator of anomalies. As pointed out in \cite{gong2019memorizing}, however, the AE can sometimes ``generalize'' so well that it can also reconstruct the abnormal inputs well. On the other hand, they are often sensitive to typical image degradations such as blur and noise and might end up identifying good samples as defective. 

\begin{figure*}
    \centering
	\includegraphics[width=0.85\textwidth]{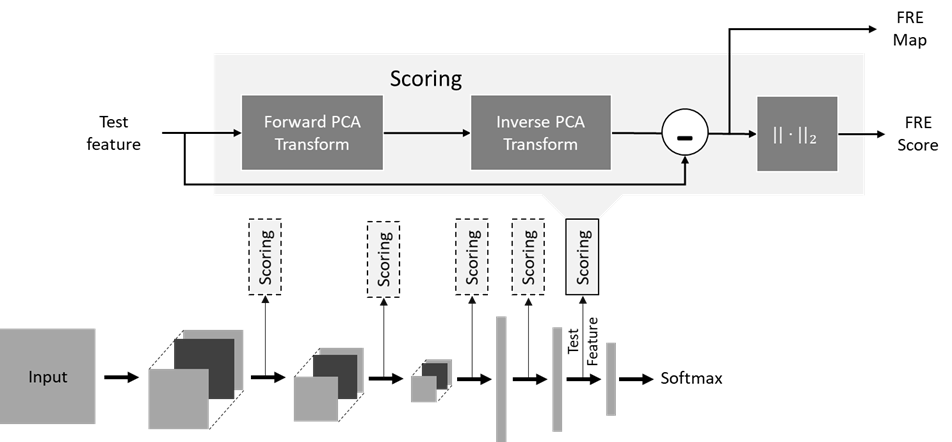}
    \caption{Block diagram show complete pipeline of our proposed approach.}
    \label{fig:pipeline}
\end{figure*}

Methods that rely on detecting anomalies in the feature space of DNNs rather than reconstruction-error in the pixel space have thus been found to be more effective and robust. Within these methods, there are one-class classification methods \cite{ruff2018deep, chalapathy2018anomaly}, and methods based on deep generative models \cite{ganomaly, differnet}, both of which are trained using defect-free data in order to model the distribution of good samples.  

More recently, excellent results have been obtained by using DNNs pretrained on a large dataset such as ImageNet and then building anomaly models on the feature spaces of such DNNs. Not only do these methods avoid DNN training (often a cumbersome process), but also are able to leverage a rich space of feature representations learnt from a larger, more diverse dataset. These include methods such as \cite{defard2021padim, cohen2020sub, roth2021towards} that model clusters or distributions on a multi-level pyramid of deep-features. 
These methods show impressive results to varying degrees, but still incur significant computational and storage complexity in building anomaly models on top of the pretrained representations.

Additionally, there is oftentimes a need to not only identify which image samples are anomalous, but also to determine the location of the defective pixels within the image for further diagnostic and remedy purposes. This is a much more challenging segmentation problem also referred to as anomaly localization. Several of the methods listed above perform both anomaly detection and anomaly localization.

We present here a method for anomaly detection and localization that combines the best aspects of the methods reveiwed above: it is based on learning low-dimensional representations from high-dimensional data (similar to deep AEs); this modeling is performed, however, in the feature space rather than the pixel space; and finally, it leverages pretrained networks thereby avoiding any costly training. 
Motivated by the \emph{manifold hypothesis}, which posits that real-world high-dimensional data lie on low-dimensional manifolds, our method involves mapping the high-dimensional deep features of a DNN into an appropriate low-dimensional subspace.
We show that the \emph{feature reconstruction error} (FRE), which is the difference between the original feature in the high-dimensional space and the pre-image of its low-dimensional reduced embedding, is highly effective for both anomaly detection and pixel level anomaly localization. This circumvents the need to perform any subsequent processing in the reduced feature space, thereby greatly simplifying the procedure both during modeling and inference. We show that this approach achieves state-of-the-art anomaly detection and segmentation performance (section \ref{sec:experiments}). Importantly, our method has significantly lower complexity compared to other methods and can be run at very high frame-rates on a CPU-only system. This makes it very attractive for deployment in real-world industrial usages on low-cost edge platforms without requiring investment into expensive discrete GPUs. Furthermore, the approach does not modify the network's parameters, which is a significant advantage for trained networks already deployed.




\paragraph{Contributions:}
In summary, we present a fast, principled approach to unsupervised image-level anomaly detection and pixel-level anomaly segmentation. We show, by means of exhaustive experimentation, that it achieves state-of-the-art performance on both tasks on industrial defect detection datasets. Finally, we demonstrate that our method can be trained in a matter of seconds and can be run at very high-frame rates with only a CPU, making it a very attractive and practical option for real-world deployment.

\section{Method/Approach}
\label{sec:Method}

Consider a deep neural network (DNN) trained on an $N$-class classification problem.
For an input $\mb{x}$, let $\mb{u}\triangleq f_k(\mb{x}) \in \mathbb{R}^{C_k\times H_k\times W_k} $ denote the output feature at the $k^{th}$ intermediate layer of the network. 

\subsection{Feature reconstruction error (FRE)}
The features induced by the training dataset do not fully span the high-dimensional space in which they reside. Hence, for a training dataset of size $M$, the data-matrix $\mb{D} = [f_k(\mb{x}_1)|\dots|f_k(\mb{x}_M)] $ constructed from the features is rank deficient. Table \ref{table:data_ranks} shows how severe the rank deficiencies are in the higher-dimensional inner layers of a Resnet18 deep network used in our experiments.
Consequently, we learn a transformation $\mathcal{T}$ that maps the high-dimensional features onto an appropriate subspace, $\mathcal{T}: \mathcal{H} \to \mathcal{L}$ with $dim(\mathcal{L}) \ll dim(\mathcal{H})$. The parameters of the inverse transformation $\mathcal{T}^{\dagger}$ are also learnt simultaneously. 

We propose to use principal component analysis (PCA) for dimensionality reduction \cite{shlens2014tutorial}. 
In this framework, $\mathcal{H}$ and  $\mathcal{L}$ are, respectively, Euclidean spaces $\mathbb{R}^d$ and $\mathbb{R}^m$, with $m \ll d$. $\mathcal{T}$ is then an orthogonal linear transformation that can be simply calculated from the singular value decomposition (SVD) \cite{golub2013matrix} of the data matrix $\mb{D}$. As the mapping from the high-dimensional feature space to the lower-dimensional reduced subspace is non-injective, there isn't a uniquely defined inverse image (pre-image) for a reduced feature, i.e. $\mathcal{T}^{\dagger}$ is not uniquely defined. A common practice is to use the Moore-Penrose pseudo-inverse of the forward transformation \cite{golub2013matrix} as the inverse. 
While we propose using PCA because of simplicity, linearity and performance, our approach can employ any dimensionality reduction or manifold learning technique that provides an explicit mapping function for new data points. 

During inference, the transformation $\mathcal{T}$ is applied to a test feature, $\mb{u}\triangleq f_k(\mb{x}) \in \mathbb{R}^{C_k\times H_k\times W_k} $ to obtain its reduced-dimension embedding. 
In practice, the output feature $\mb{u}$ is vectorized prior to dimensionality reduction. This reduced embedding is inverse-transformed into the original space by appplying $\mathcal{T}^{\dagger}$ and a \textit{feature reconstruction error} (FRE) is calculated as the difference between the original and reconstructed vectors, as given by
\begin{equation}
\label{eq:FRE}
    FRE(\mb{u}, \mathcal{T}) = \mb{e} \triangleq \mb{u}-(\mathcal{T}^{\dagger} \circ \mathcal{T})\mb{u}.
\end{equation}
Note that $FRE \in \mathbb{R}^{C_k\times H_k\times W_k} $ has the same dimension as the feature tensor $\mb{u}$, owing to the inverse transformation $\mathcal{T}^{\dagger}$. 
The FRE detection score, then defined as
\begin{equation}
\label{eq:FREscore}
    FRE_{score}(\mb{u}, \mathcal{T}) = \|\mb{e}\|_2,
\end{equation}
is highly effective at discriminating between in-distribution (normal) and out-of-distribution (abnormal) samples. The intuition behind FRE is that OOD samples will lie outside the subspace of ID samples and will hence result in higher FRE scores. 

\begin{table}
    \footnotesize
	\caption{Feature dimensions and data-matrix ranks for Resnet18 trained on CIFAR10}
	\label{table:data_ranks}
	\centering
	\begin{tabular}{cccc}
		\toprule
		\textbf{Layer} & Layer 0 & Layer 1 & Layer 2\\
		\cmidrule(r){1-1}\cmidrule(r){2-4}
		Dimension & 512 & 256 & 512 \\
		Rank & 85 & 255 & 478 \\
		With $99.5\%$ PCA & 29 & 239 & 463 \\
		\bottomrule
	\end{tabular}
\end{table}

We next show how our method can also be used to derive pixel anomaly maps for anomaly localization. To obtain these, we start with the FRE, $\mb{e}$, from Eq. (\ref{eq:FRE}) and reshape it to match the size of the feature tensor,  i.e. $C_k$ channels each of size $H_k \times W_k$. We then perform channel-wise averaging at all locations to produce a single-channel FRE anomaly map of size $H_k \times W_k$, i.e.
\begin{equation*}
    \mb{M}_k(i,j) = \frac{1}{C_k}\sum_{c=1}^{C_k} \mb{e}(c,i,j)
\end{equation*}
Finally, we resize this map to the dimension of the input image $H \times W$ in order to produce the final anomaly map $\mb{M}$. This anomaly map is a heatmap of the same resolution as the input image, with higher intensity regions corresponding to anomalies. Taken together, the FRE score and anomaly map derived from the intermediate features of a DNN suffice to perform both anomaly detection and pixel-level anomaly localization (segmentation). A block diagram of the complete end-to-end pipeline is shown in Fig. \ref{fig:pipeline}.

\paragraph{Layer selection:} Our technique can be applied to the features from any intermediate layer of a DNN. It is expected that features from different layers would have different characteristics which could lead to different performance outcomes. It is well-known, for instance, that layers close to the input learn spatial representations that are good for localization, while downstream layers closer to the output learn representations that are semantically richer. In Section \ref{sec:experiments}, we investigate the impact of the choice of layers on the quality of anomaly localization by running our method on layers positioned at different points within the  network. As will be seen, the best anomaly localization performance is obtained at layers positioned around the center of the network, where the features have a good balance of spatial and semantic characteristics.

However, we can further extend and generalize our method by not restricting the feature modeling to a single layer but instead  generating FRE maps at multiple layers and combining these into a single map. In Section \ref{sec:experiments}, we study the improvement that can be obtained by such combination on a variety of pretrained DNN models (backbones)


\subsection{Computational Complexity}
Existing state-of-the-art methods for anomaly detection and localization involve significant complexity during training or inference. \cite{differnet, ganomaly} use deep generative models (GANs or normalizing flows) to model the distribution of normal samples, which require expensive training. While \cite{cohen2020sub, defard2021padim} mititgate training complexity by using pretrained models, they both involve modeling clusters or probability distributions on a multi-level pyramid of deep-features. 
\cite{roth2021towards} reduces the memory storage requirements of feature-pyramid based methods but uses greedy coreset subsampling on the stored feature banks to accomplish this, which is known to be a computationally involved process (NP-hard). By contrast, our method does not involve training a new model of any kind (discriminative or generative), operates on features from a single layer of the deep-network (instead of a feature pyramid), and does not involve any complex probabilistic modeling. The only processing in our method (other than a DNN forward pass) are forward and inverse PCA transformations, which are simply matrix multiplication operations. The computational overhead compared to a forward pass of the CNN backbone is very low with the result that our method can be run at very high frame rates \emph{without requiring any discrete GPUs}. This makes it very attractive to deploy in real-world industrial usages.

\section{Experiments}
\label{sec:experiments}
\begin{table}
    \scriptsize
	\caption{Anomaly Detection AUROC on MVTec dataset \\{\footnotesize $^\star$ Methods not reporting class-itemized results. For PaDiM, results were recreated using \emph{anomalib}\cite{anomalib}}}
	\label{table:mvtec_AD}
	\centering
	\begin{tabular}{ccccccc}
	
		\toprule
		 \multirow{2}{*}{Category} & GAN- & Differ & SPADE$^\star$ & PaDim$^\star$ & Patch & FRE \\
		  & -omaly & Net &  &  & Core & (Ours) \\		
    \midrule
    Carpet & 69.9 & 92.9 & - & 99.5 & 98.7 & 100 \\
		\midrule
    Grid & 70.8 & 84.0 & - & 94.2 & 98.2 & 95.8 \\
		\midrule
    Leather & 84.2 & 97.1 & - & 100 & 100 & 100 \\
		\midrule
    Tile & 79.4 & 99.4 & - & 97.4 & 98.7 & 97.8 \\
		\midrule
    Wood & 83.4 & 99.8 & - & 99.3 & 99.2 & 99.4 \\
		\midrule
    Bottle & 89.2 & 99.0 & - & 99.9 & 100 & 100 \\
		\midrule
    Cable & 75.7 & 95.9 & - & 87.8 & 99.5 & 99.3 \\
		\midrule
    Capsule & 73.2 & 86.9 & - & 92.7 & 98.1 & 99.4 \\
		\midrule
    Hazelnut & 78.5 & 99.3 & - & 96.4 & 100 & 99.8 \\
		\midrule
    Metal Nut & 70.0 & 96.1 & - & 98.9 & 100 & 96.9 \\
		\midrule
    Pill & 74.3 & 88.8 & - & 93.9 & 96.6 & 95.7 \\
		\midrule
    Screw & 74.6 & 96.3 & - & 84.5 & 98.1 & 97.5 \\
		\midrule
    Toothbrush & 65.3 & 98.6 & - & 94.2 & 100 & 99.4 \\
		\midrule
    Transistor & 79.2 & 91.1 & - & 97.6 & 100 & 98.6 \\
		\midrule
    Zipper & 74.5 & 95.1 & - & 88.2 & 99.4 & 96.4 \\
		\midrule
    \textbf{Average} & \textbf{76.2} & \textbf{94.9} & \textbf{85.5} & \textbf{97.9} & \textbf{99.1} & \textbf{98.4} \\
		\bottomrule	
	\end{tabular}
\end{table}

\begin{table}
    \scriptsize
	\caption{Anomaly Detection AUROC on Magnetic Tile}
	\label{table:mtd_AD}
	\centering
	\begin{tabular}{ccccc}
	
		\toprule
		\midrule
		 GANomaly & I-NN & DifferNet & PatchCore & FRE (Ours) \\
 		\\
		 
		\midrule
     \textbf{76.6} & \textbf{80.0} & \textbf{97.7} & \textbf{97.9}  & \textbf{99.2} \\
		\bottomrule
	\end{tabular}
\end{table}

	

\begin{table}
    \scriptsize
	\caption{Anomaly Detection AUROC across backbones (MVTec)}
	\label{table:mvtec_AD2}
	\centering
	\begin{tabular}{cccc}
	
		\toprule
		 ResNet18 & ResNet50 & VGG16 & EfficientNet-B5 \\
		\midrule
      \textbf{94.1} & \textbf{94.2} & \textbf{91.1} & \textbf{98.4} \\
		\bottomrule	
	\end{tabular}
\end{table}

\begin{figure*}
	\centering
	\includegraphics[width=0.16\textwidth]{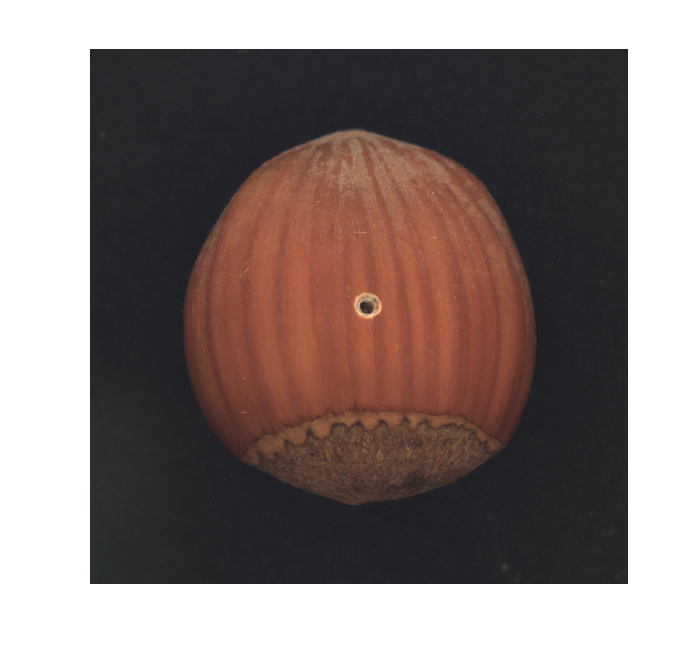}
	\includegraphics[width=0.16\textwidth]{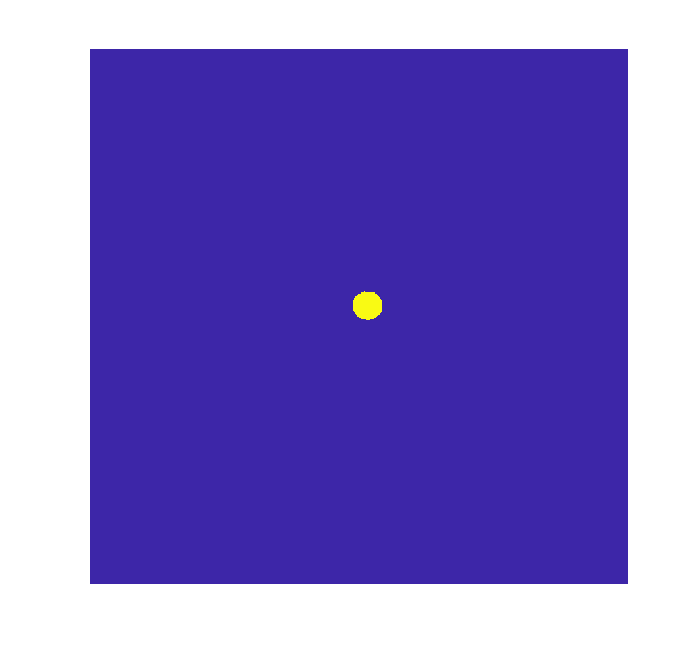}
	\includegraphics[width=0.16\textwidth]{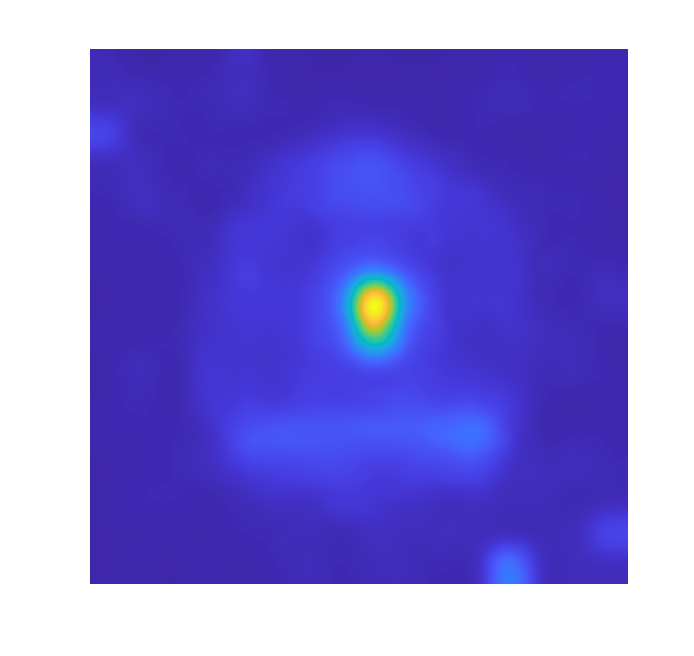}
	\includegraphics[width=0.16\textwidth]{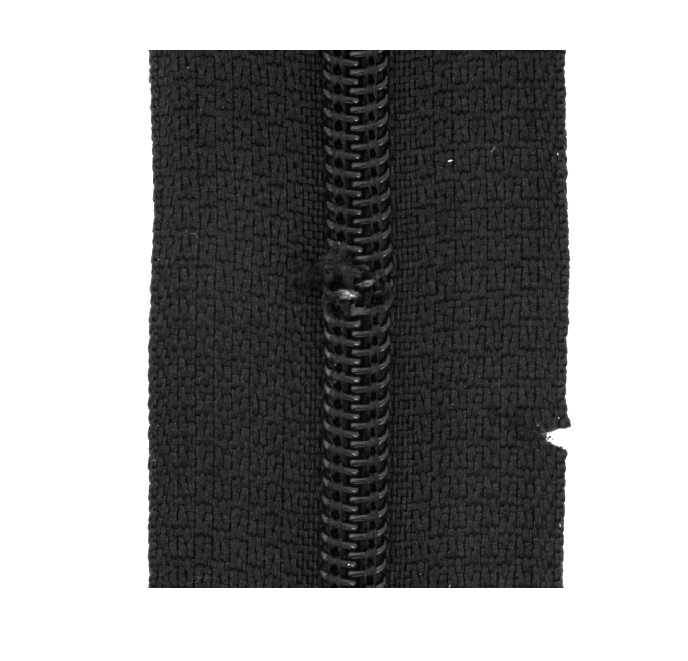}
	\includegraphics[width=0.16\textwidth]{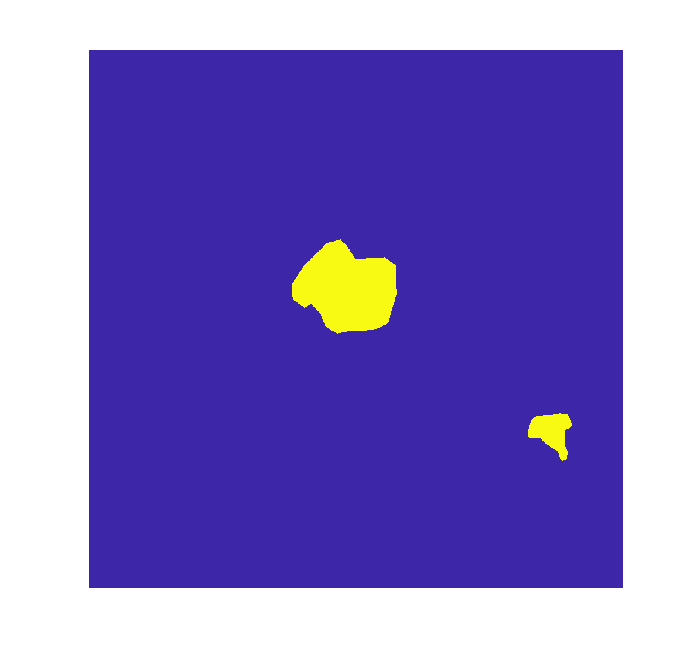}
	\includegraphics[width=0.16\textwidth]{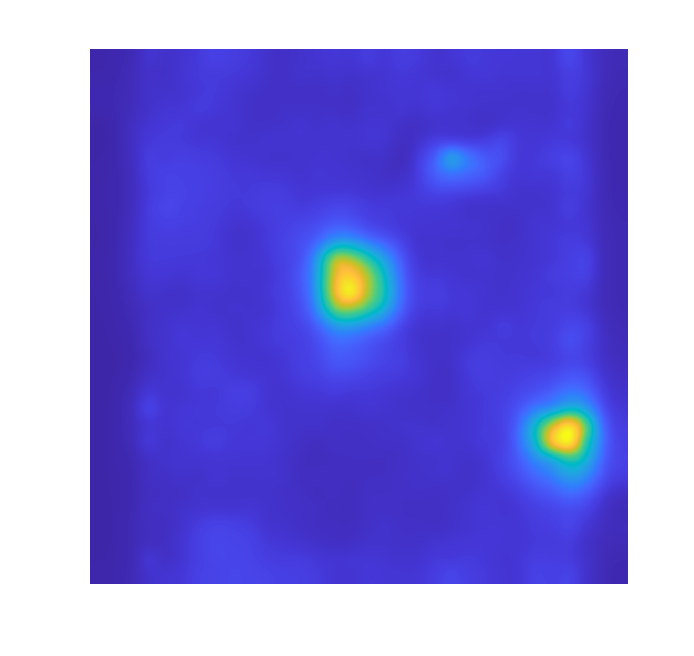}
	\includegraphics[width=0.16\textwidth]{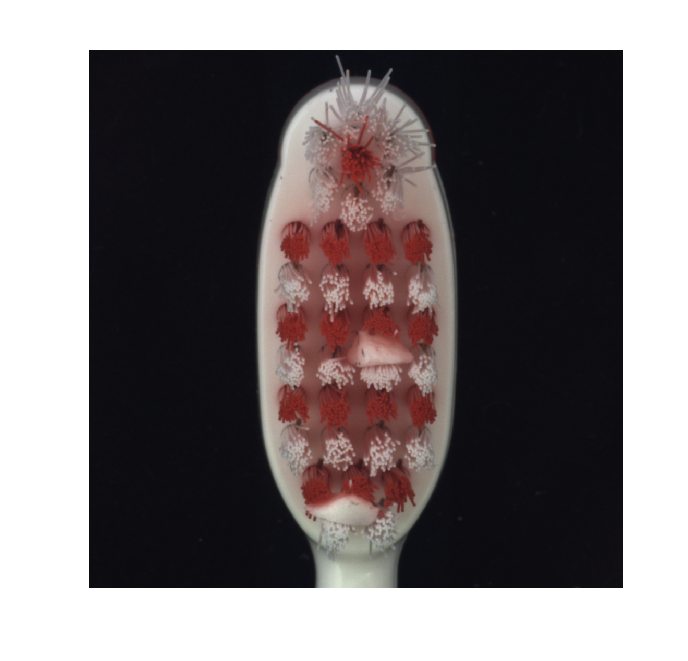}
	\includegraphics[width=0.16\textwidth]{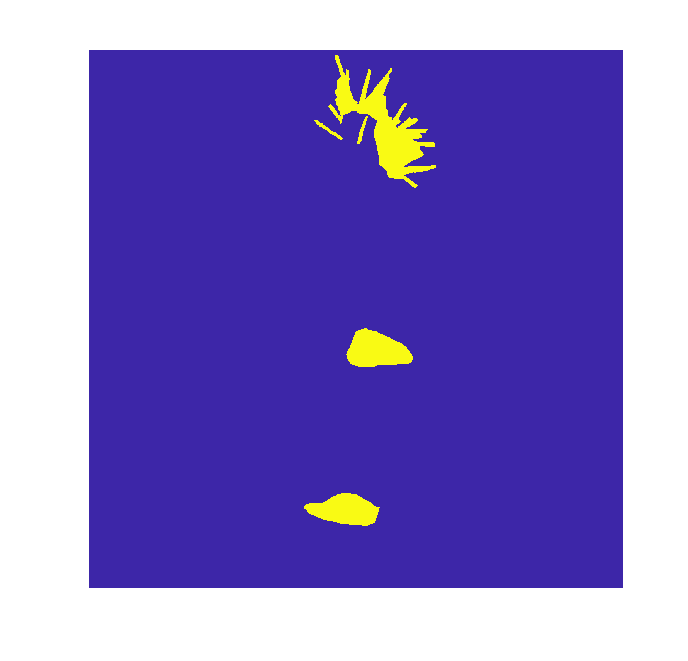}
	\includegraphics[width=0.16\textwidth]{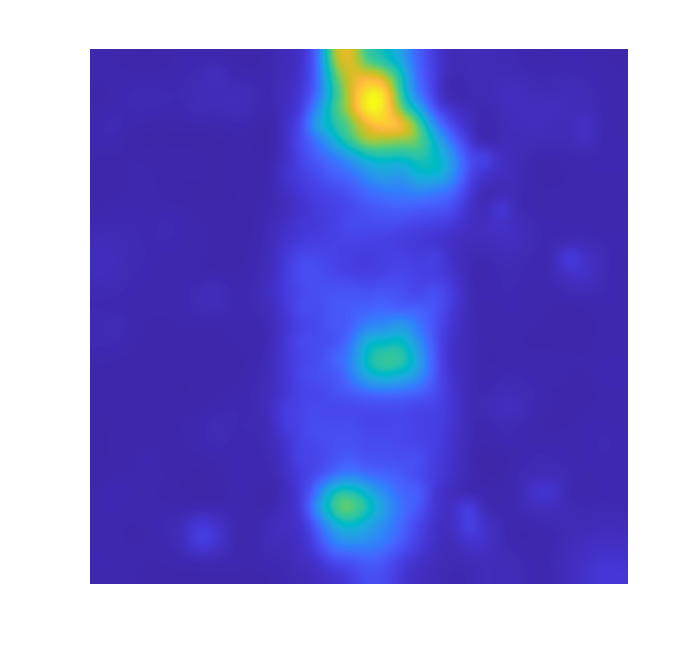}
	\includegraphics[width=0.16\textwidth]{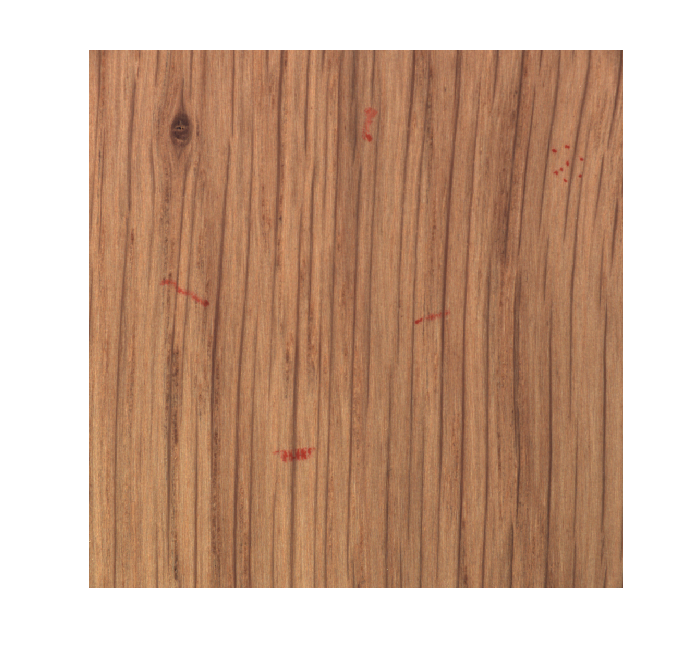}
	\includegraphics[width=0.16\textwidth]{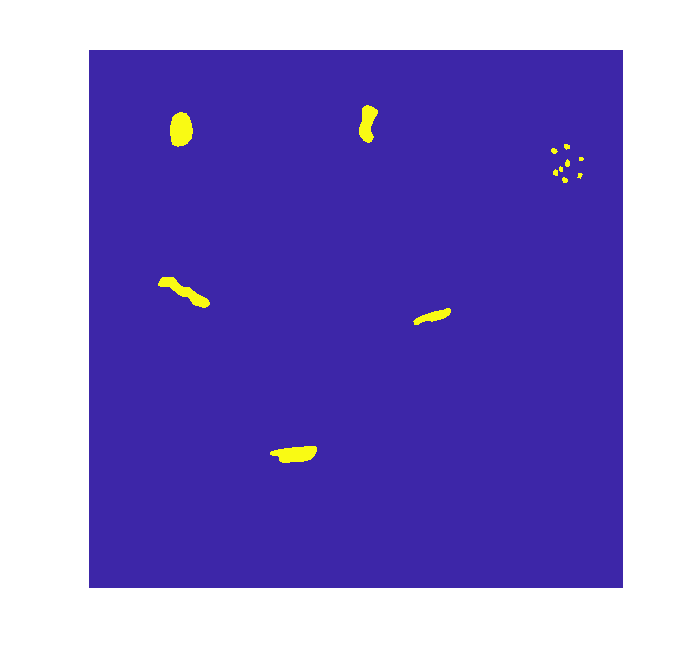}
	\includegraphics[width=0.16\textwidth]{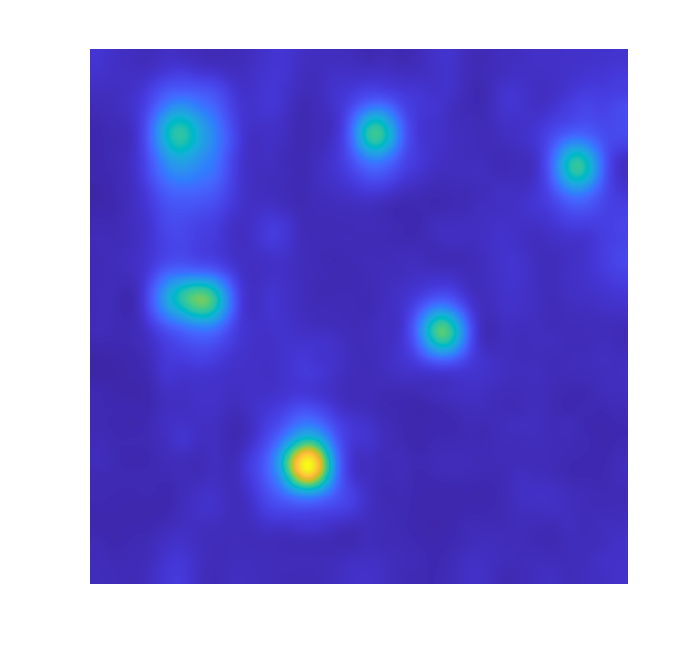}
	\includegraphics[width=0.16\textwidth]{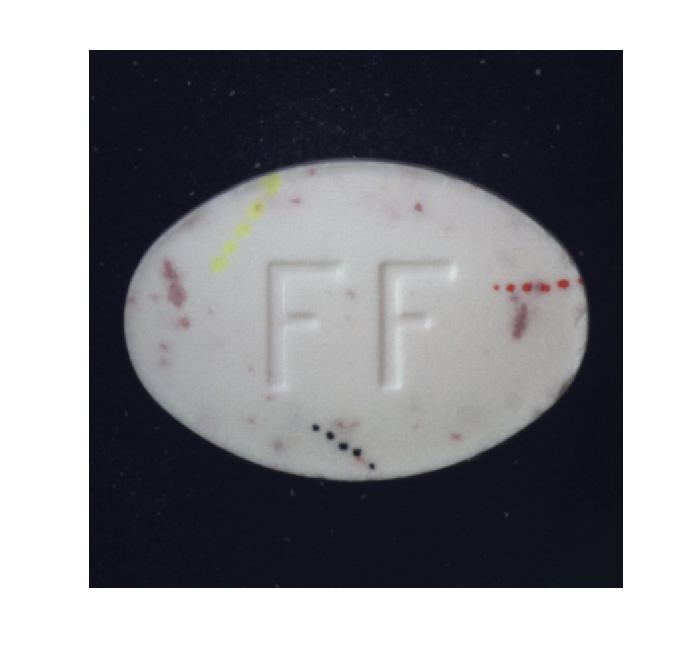}
	\includegraphics[width=0.16\textwidth]{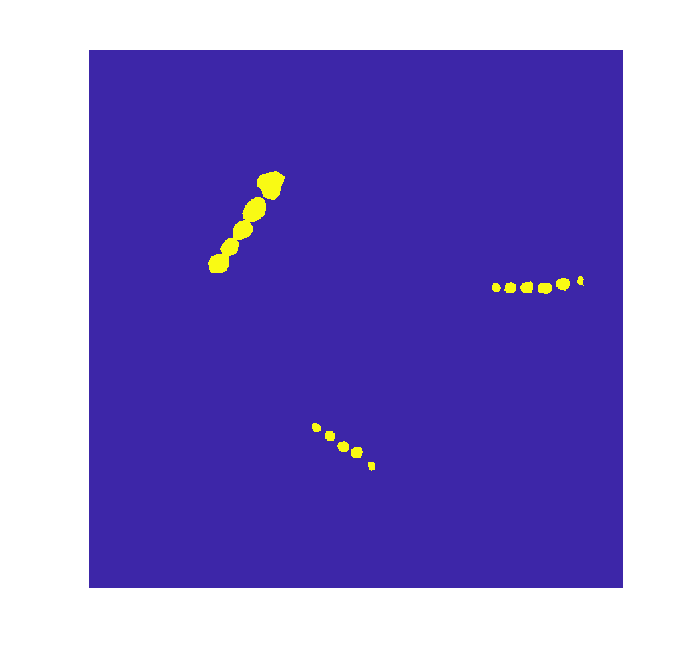}
	\includegraphics[width=0.16\textwidth]{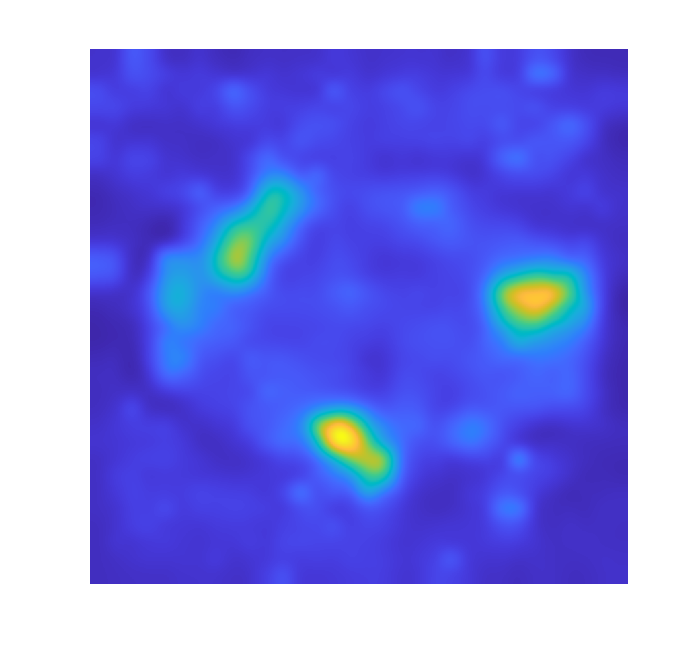}
	\includegraphics[width=0.16\textwidth]{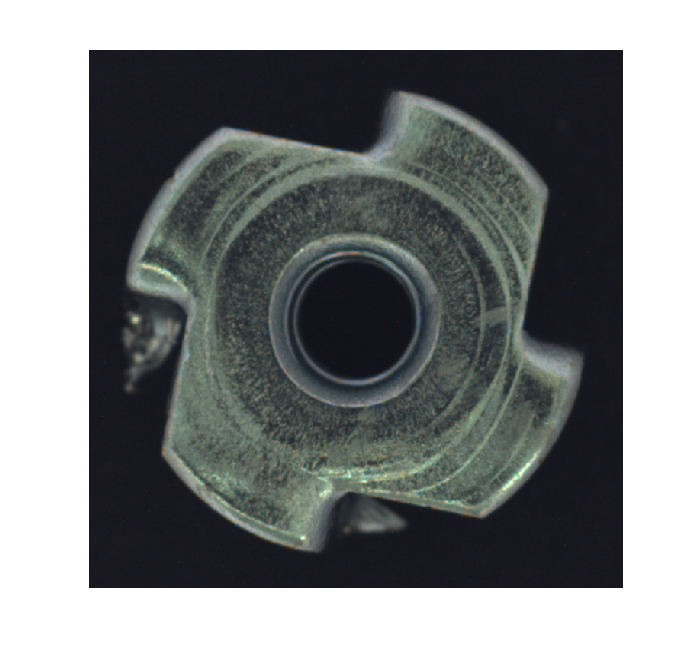}
	\includegraphics[width=0.16\textwidth]{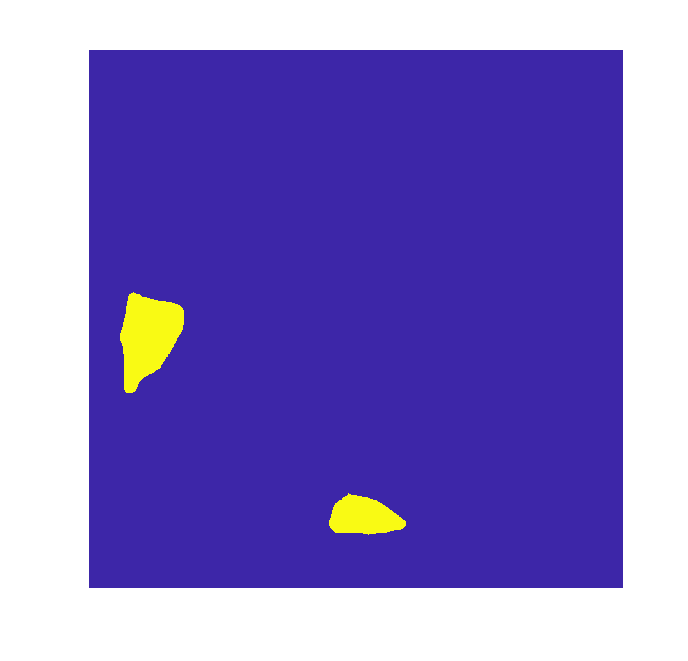}
	\includegraphics[width=0.16\textwidth]{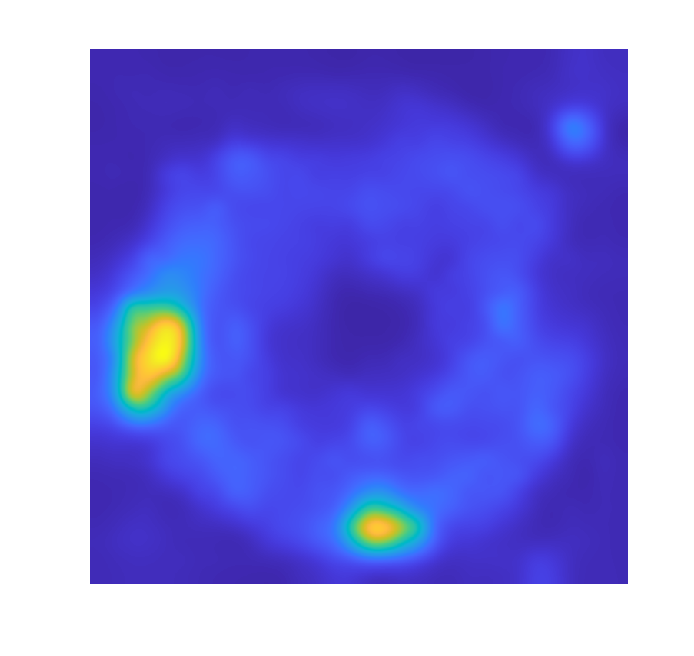}
	\includegraphics[width=0.16\textwidth]{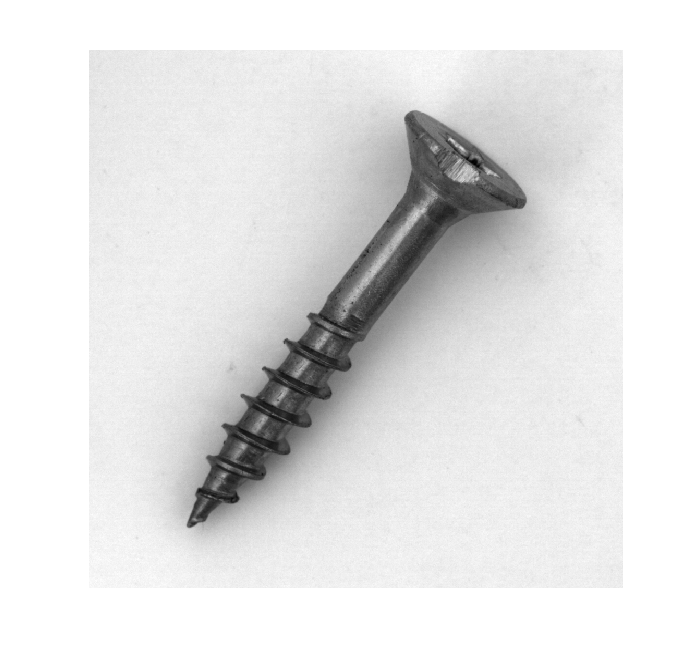}
	\includegraphics[width=0.16\textwidth]{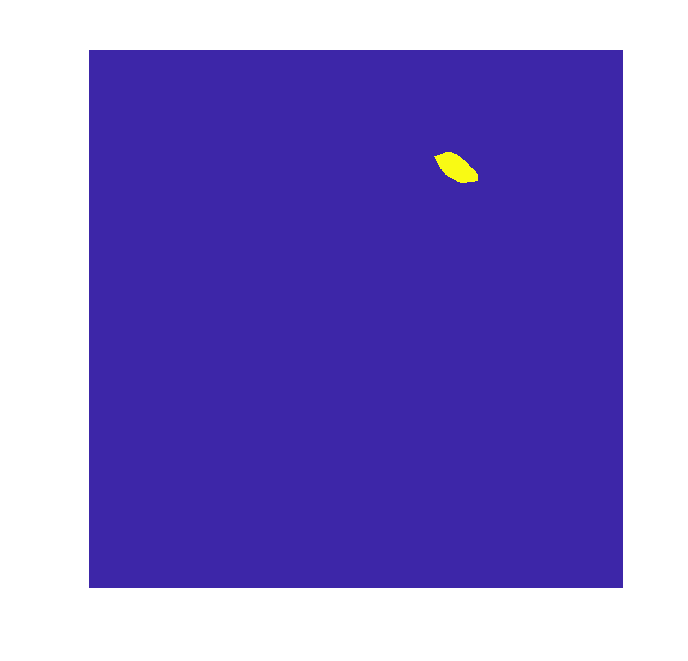}
	\includegraphics[width=0.16\textwidth]{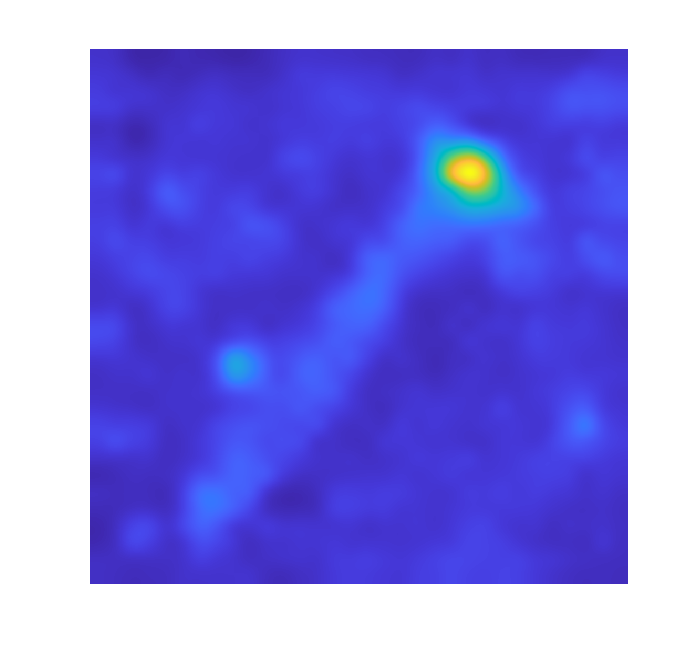}
	\includegraphics[width=0.16\textwidth]{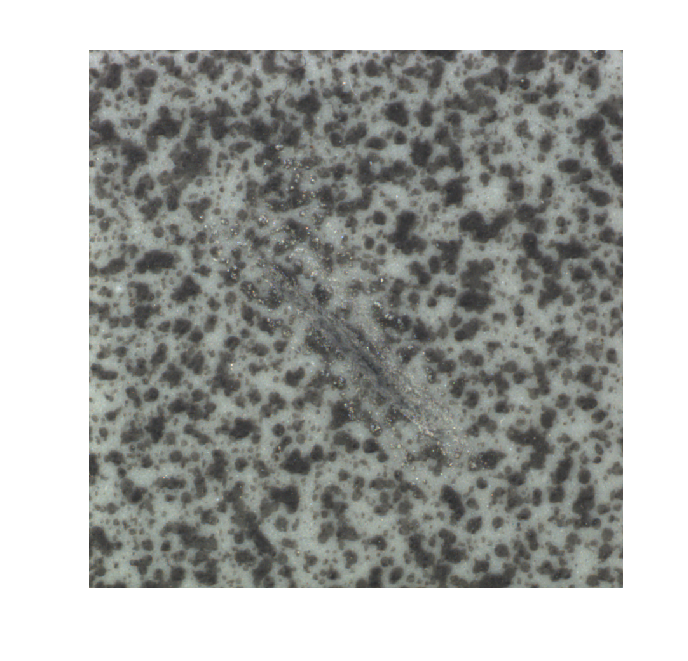}
	\includegraphics[width=0.16\textwidth]{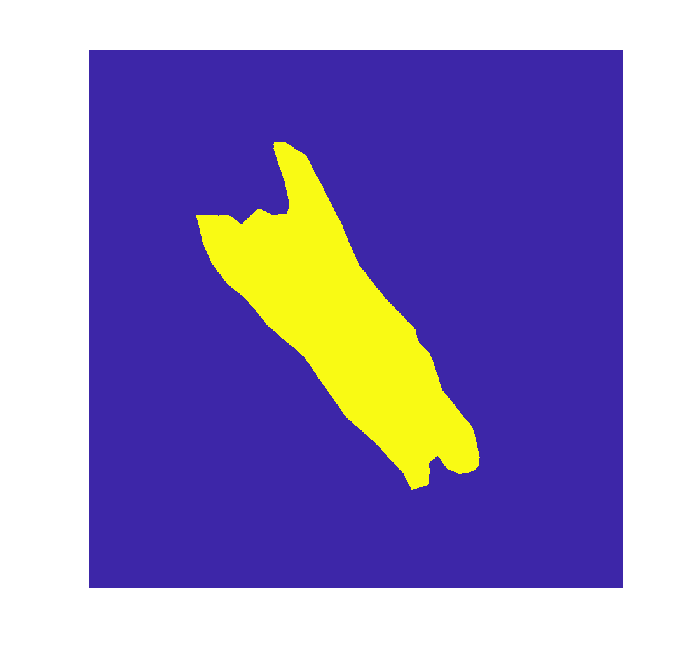}
	\includegraphics[width=0.16\textwidth]{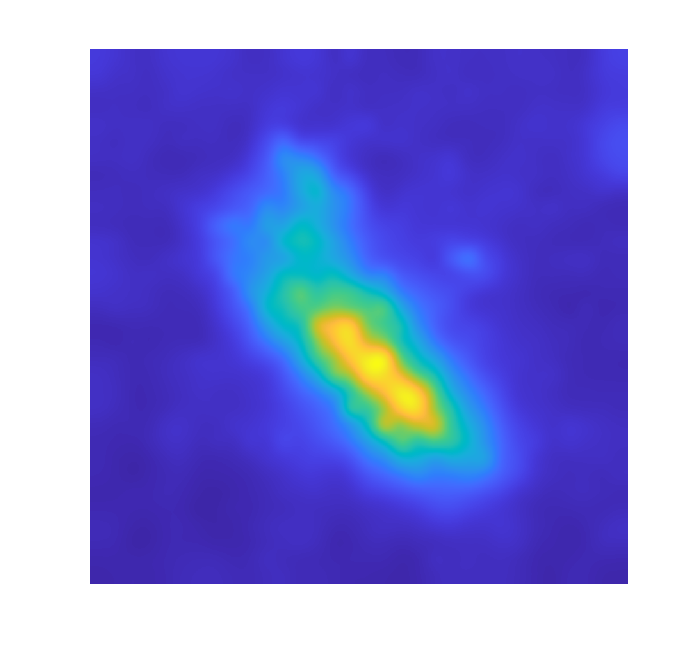}
	\includegraphics[width=0.16\textwidth]{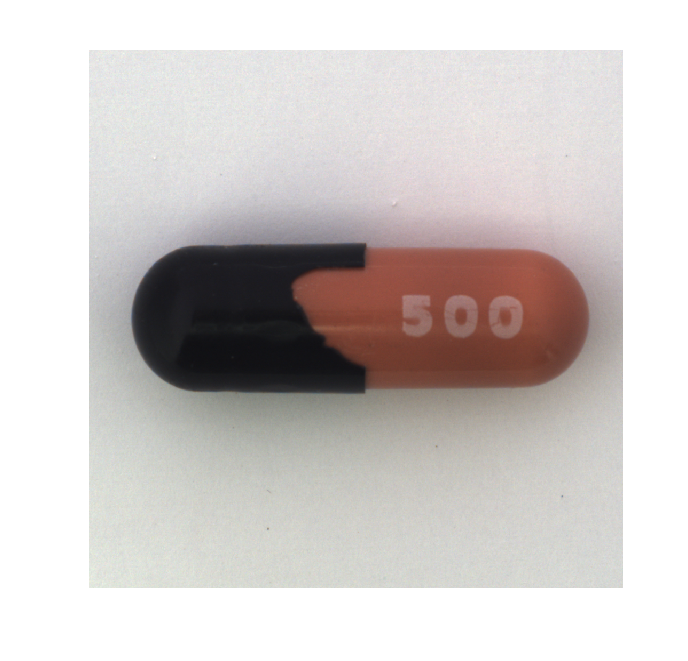}
	\includegraphics[width=0.16\textwidth]{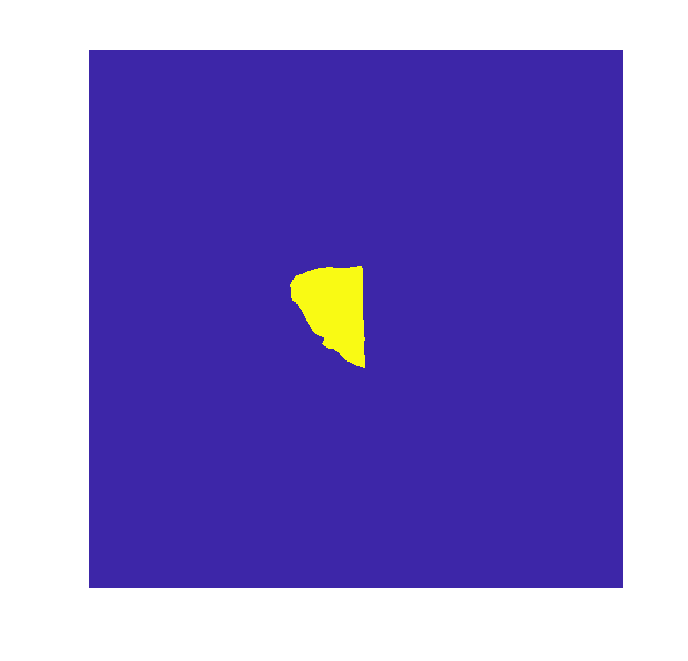}
	\includegraphics[width=0.16\textwidth]{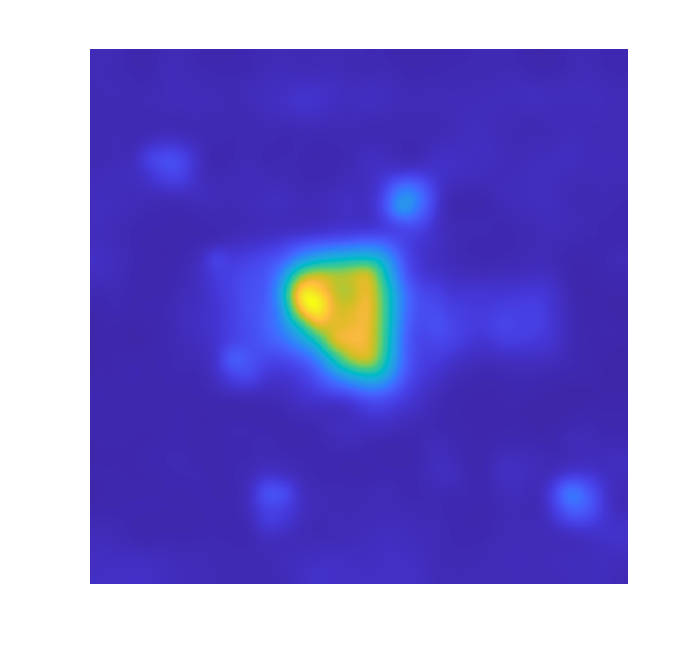}
	\includegraphics[width=0.16\textwidth]{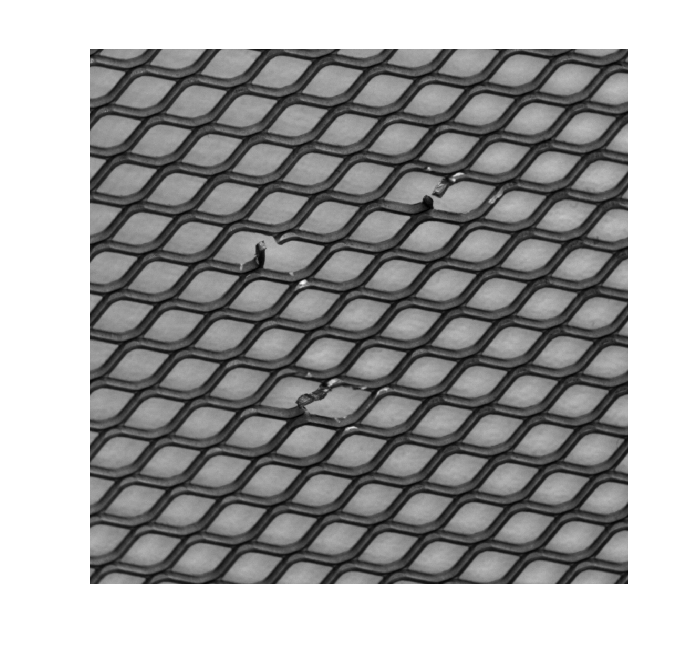}
	\includegraphics[width=0.16\textwidth]{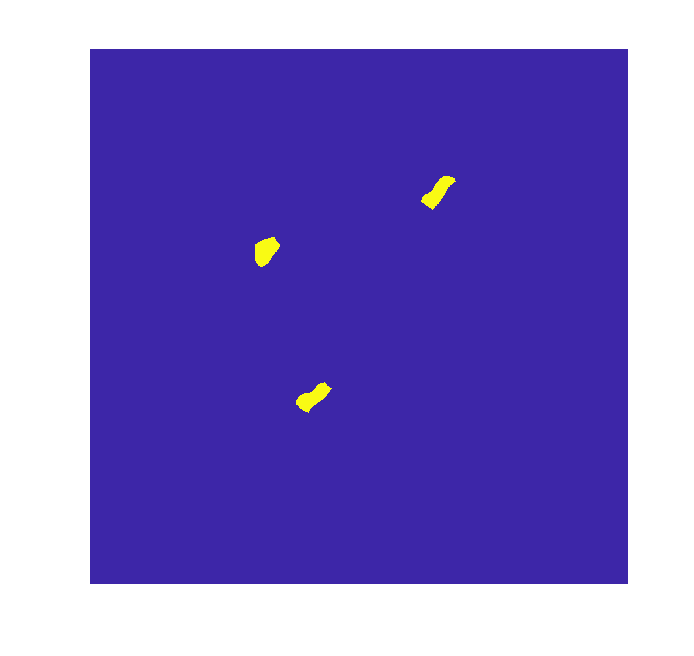}
	\includegraphics[width=0.16\textwidth]{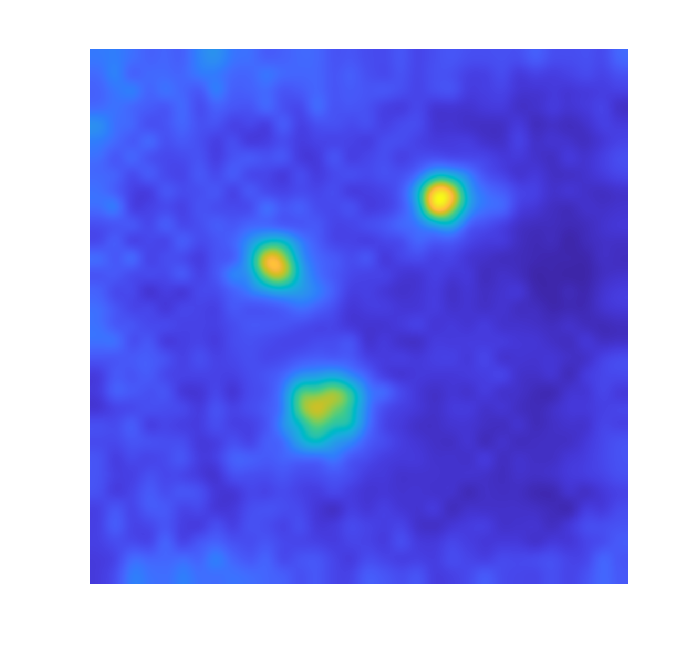}

	\caption{From left to right, each set of three images comprises: Original image, ground truth segmentation mask, anomaly heatmap using FRE (our method).}
	\label{fig:sample_outputs}
\end{figure*}

\begin{figure*}
	\centering
	\includegraphics[width=0.12\textwidth]{cherry2/input_042.png}
	\includegraphics[width=0.12\textwidth]{cherry2/gt_042.png}
	\includegraphics[width=0.12\textwidth]{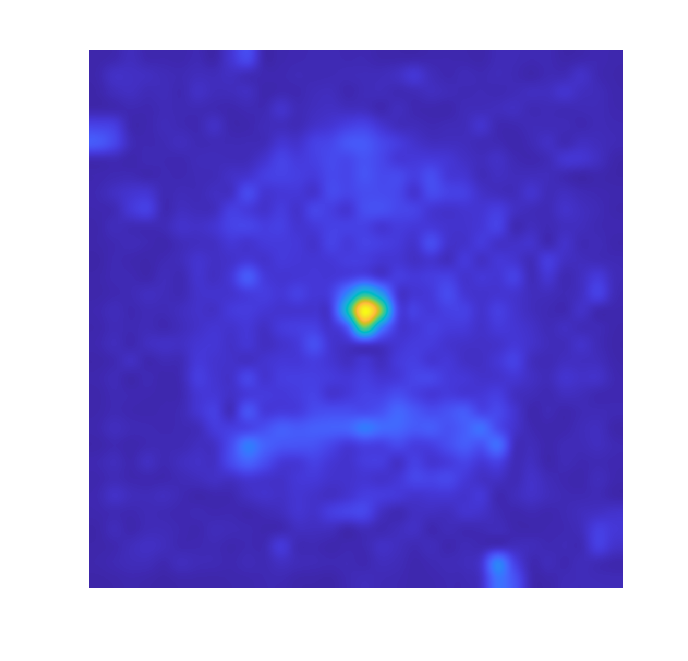}
	\includegraphics[width=0.12\textwidth]{cherry2/042_combined.png}
	\includegraphics[width=0.12\textwidth]{cherry/input_029.png}
	\includegraphics[width=0.12\textwidth]{cherry/gt_029.png}
	\includegraphics[width=0.12\textwidth]{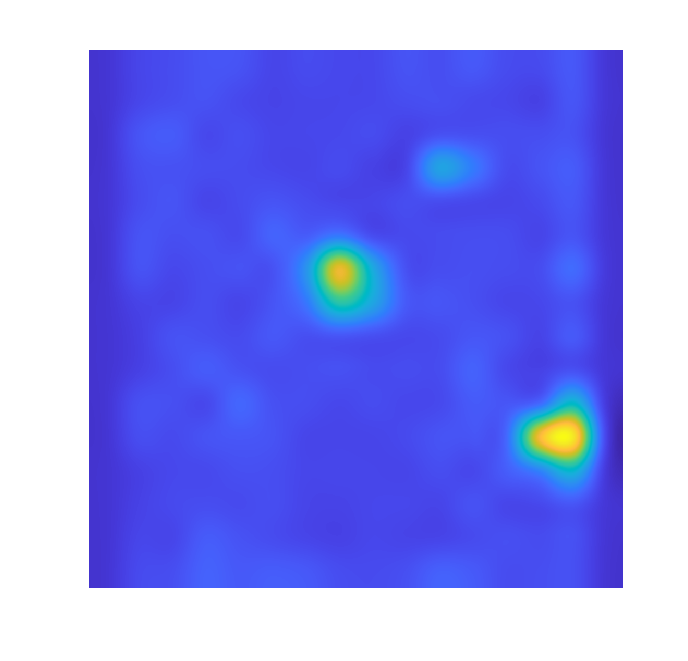}
	\includegraphics[width=0.12\textwidth]{cherry2/029_combined.png}
	\includegraphics[width=0.12\textwidth]{cherry/_input_014.png}
	\includegraphics[width=0.12\textwidth]{cherry/_gt_014.png}
	\includegraphics[width=0.12\textwidth]{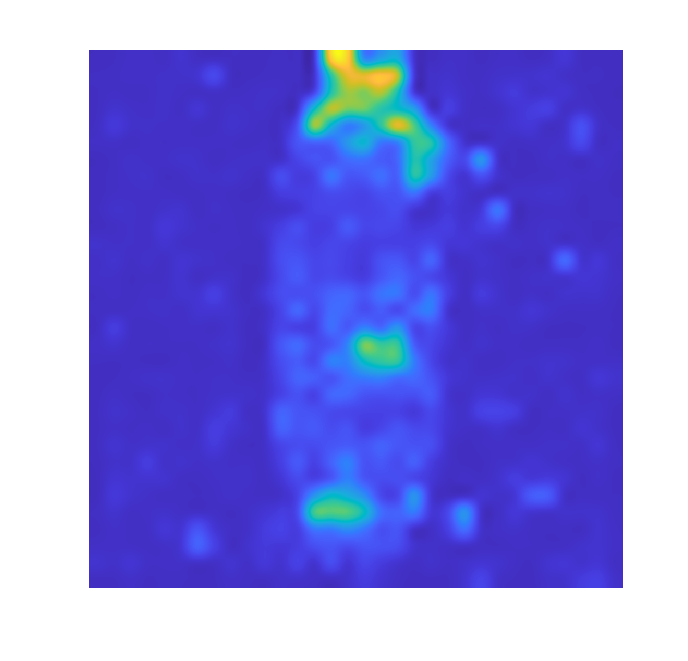}
	\includegraphics[width=0.12\textwidth]{cherry2/__014_combined.png}
	\includegraphics[width=0.12\textwidth]{cherry/input_013.png}
	\includegraphics[width=0.12\textwidth]{cherry/gt_013.png}
	\includegraphics[width=0.12\textwidth]{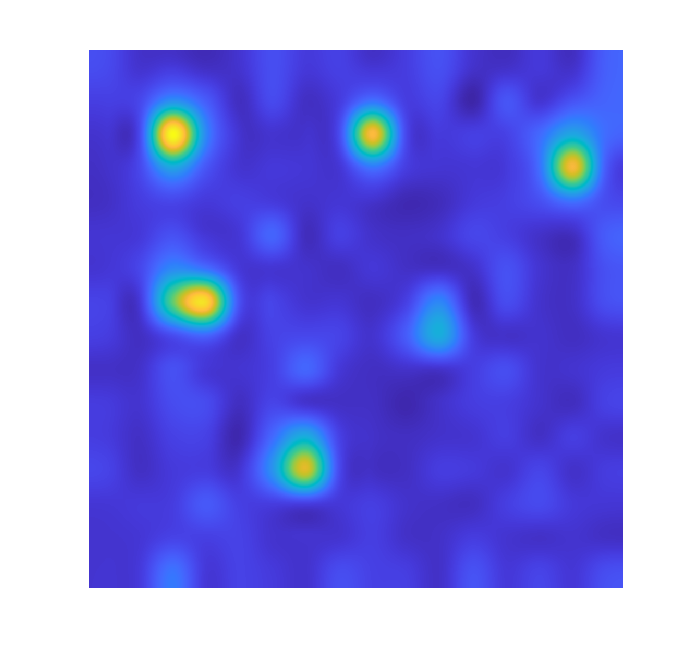}
	\includegraphics[width=0.12\textwidth]{cherry2/013_combined.png}
	\includegraphics[width=0.12\textwidth]{cherry/input_039.png}
	\includegraphics[width=0.12\textwidth]{cherry/gt_039.png}
	\includegraphics[width=0.12\textwidth]{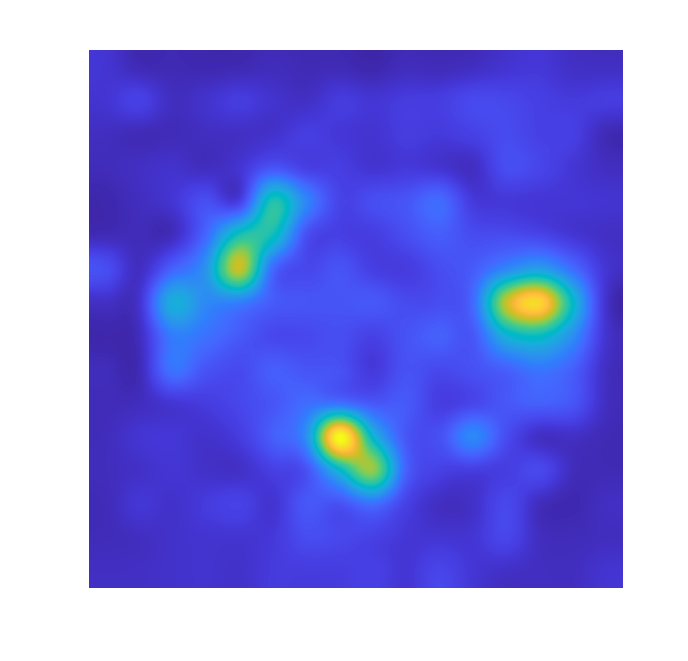}
	\includegraphics[width=0.12\textwidth]{cherry2/039_combined.png}
	\includegraphics[width=0.12\textwidth]{cherry/input_005.png}
	\includegraphics[width=0.12\textwidth]{cherry/gt_005.png}
	\includegraphics[width=0.12\textwidth]{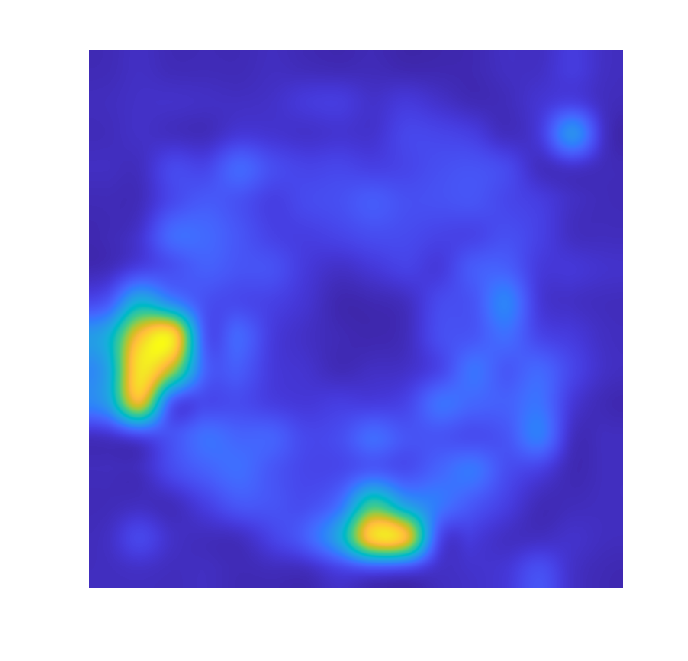}
	\includegraphics[width=0.12\textwidth]{cherry2/005_combined.png}
	\includegraphics[width=0.12\textwidth]{cherry/input_028.png}
	\includegraphics[width=0.12\textwidth]{cherry/gt_028.png}
	\includegraphics[width=0.12\textwidth]{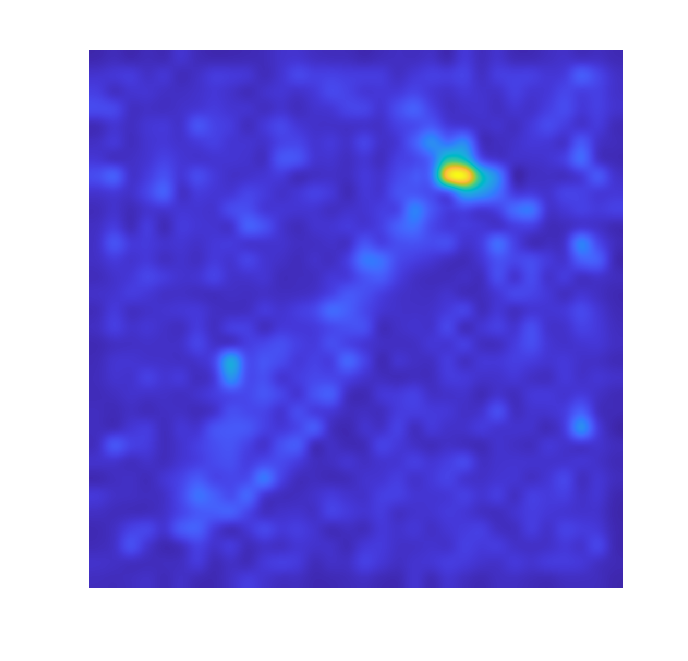}
	\includegraphics[width=0.12\textwidth]{cherry2/028_combined.png}
	\includegraphics[width=0.12\textwidth]{cherry/_input_071.png}
	\includegraphics[width=0.12\textwidth]{cherry/_gt_071.png}
	\includegraphics[width=0.12\textwidth]{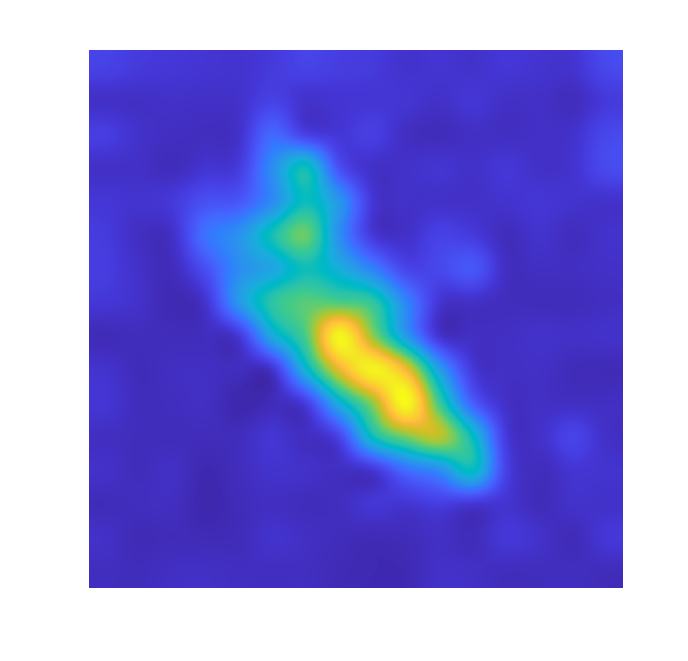}
	\includegraphics[width=0.12\textwidth]{cherry2/__071_combined.png}
	\includegraphics[width=0.12\textwidth]{cherry/input_010.png}
	\includegraphics[width=0.12\textwidth]{cherry/gt_010.png}
	\includegraphics[width=0.12\textwidth]{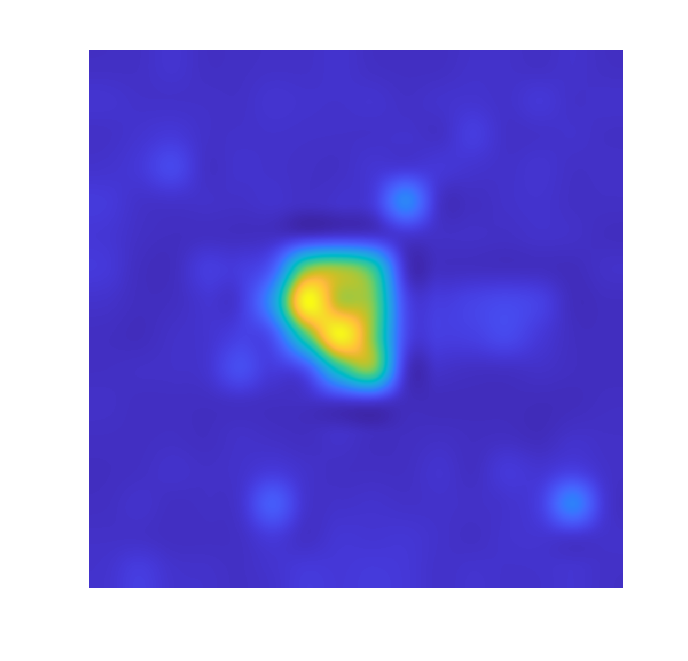}
	\includegraphics[width=0.12\textwidth]{cherry2/010_combined.png}
	\includegraphics[width=0.12\textwidth]{cherry2/input_018.png}
	\includegraphics[width=0.12\textwidth]{cherry2/gt_018.png}
	\includegraphics[width=0.12\textwidth]{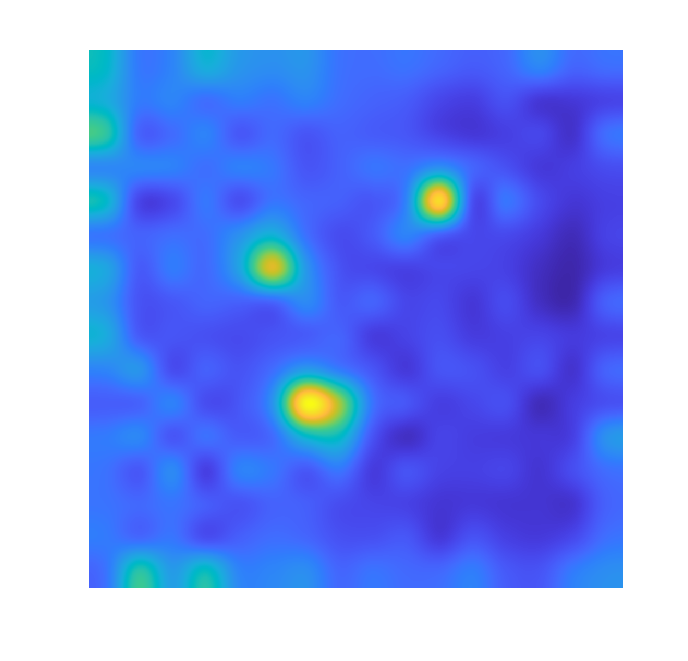}
	\includegraphics[width=0.12\textwidth]{cherry2/018_combined.png}

	\caption{\textbf{Single-layer vs 3-layer heatmaps.} From left to right, each set of four images comprises: Original image, ground truth segmentation mask, anomaly heatmap using FRE (our method) from a single layer, anomaly heatmap using FRE from three layers. Notice that 3-layer maps appear somewhat cleaner.}
	\label{fig:sample_outputs}
\end{figure*}

We perform a detailed study of our method, experimenting with several pretrained DNN models, and comparing our results against multiple benchmarks using objective metrics of quality and performance. We rely on two popular datasets for industrial anomaly detection that have been extensively used and cited in the literature: the Magnetic Tile Defect (MTD) dataset \cite{huang2020surface} and the MVTec AD dataset \cite{bergmann2019mvtec} which is an aggregation of 15 different datasets across a variety of objects, textures and defect artifacts. Similar to some of the other methods in the literature, we resize the input images to $256\times 256$ prior to applying our methodology. 

\paragraph{Comparison against existing methods:}
In the image-level detection task, the FRE score (Eq. (\ref{eq:FREscore})) is used to distinguish between in-distribution and out-of-distribution data. This effectively creates a binary classifier, whose performance is characterized by the receiver operating characteristics (ROC) curve and we report the area under the ROC curve (AUROC) in Tables ~\ref{table:mvtec_AD}, ~\ref{table:mtd_AD} and ~\ref{table:mvtec_AD2}. In the segmentation task, we compare the FRE anomaly map against the ground truth maps using pixelwise AUROC and the PRO metric \cite{bergmann_student}. The PRO metric is well-known to better measure the segmentation quality, especially as it relates to small objects. For both image-level detection and segmentation, we benchmark against a variety of state-of-the-art anomaly detection techniques \cite{padim, patchcore, spade}. The results are presented in Tables \ref{tab:auc_benchmarks} and \ref{tab:pro_benchmarks}. We see that our method attains excellent results, competitive or exceeding the state-of-the-art.

\paragraph{Choice of backbones:} 
We investigate the effectiveness of our approach by applying it to a variety of commonly used pretrained backbone networks which include VGG16, Resnet18, Resnet50, and EfficientNet-B5. 
The AUC and PRO results for all MVTEC categories are presented in Tables \ref{tab:auc_nets} and \ref{tab:pro_nets} respectively. We see that our method attains excellent results on all backbones, demonstrating that our method is not sensitive to the choice of the architecture used. $1L$ and $3L$ indicate results obtained by either using FRE maps from a single layer or by fusing FRE maps from three layers, respectively, as is explained next.


\begin{table}[]
    \centering
    \scriptsize
    \begin{tabular}{cccccccc}
    \toprule
  & CNN  & AE & \multirow{2}{*}{SPADE} & \multirow{2}{*}{PaDiM} & Patch & \multicolumn{2}{c}{FRE (Ours)} \\
  & Dict & L2 &  &  & Core  & 1L & 3L \\

 \midrule
carpet & 72 & 59 & 97.5 & 99.1 & 99 & 98.3 & 99.2 \\
grid & 59 & 90 & 93.7 & 97.3 & 98.7 & 97.3 & 98.1 \\
leather & 87 & 75 & 97.6 & 99.2 & 99.3 & 99.8 & 99.8 \\
tile & 93 & 51 & 87.4 & 94.1 & 95.6 & 95.1 & 96.5 \\
wood & 91 & 73 & 88.5 & 94.9 & 95 & 96.7 & 97.7 \\
bottle & 78 & 86 & 98.4 & 98.3 & 98.6 & 98.7 & 98.8 \\
cable & 79 & 86 & 97.2 & 96.7 & 98.4 & 97.4 & 97 \\
capsule & 84 & 88 & 99 & 98.5 & 98.8 & 98.8 & 99.1 \\
hazelnut & 72 & 95 & 99.1 & 98.2 & 98.7 & 98.7 & 99.1 \\
metal nut & 82 & 86 & 98.1 & 97.2 & 98.4 & 96.6 & 97.3 \\
pill & 68 & 85 & 96.5 & 95.7 & 97.4 & 96.9 & 97 \\
screw & 87 & 96 & 98.9 & 98.5 & 99.4 & 98.9 & 99.2 \\
toothbrush & 77 & 93 & 97.9 & 98.8 & 98.7 & 98.4 & 98.7 \\
transistor & 66 & 86 & 94.1 & 98.5 & 96.3 & 96.3 & 96.5 \\
zipper & 76 & 77 & 96.5 & 98.5 & 98.8 & 98.6 & 98.8 \\
\midrule
\textbf{Average} & \textbf{78} & \textbf{82} & \textbf{96} & \textbf{97.5} & \textbf{98.1} & \textbf{97.8} & \textbf{98.2} \\
\bottomrule
    \end{tabular}
    \caption{MVTec Anomaly Segmentation benchmark: pixel-wise AUROC}
    \label{tab:auc_benchmarks}
\end{table}

\begin{table}[]
    \centering
    \scriptsize
    \begin{tabular}{ccccccccc}
    \toprule
  & CNN  & AE & \multirow{2}{*}{SPADE} & \multirow{2}{*}{PaDiM} & Patch & \multicolumn{2}{c}{FRE (Ours)} \\
  & Dict & L2 &  &  & Core & 1L & 3L \\

 \midrule
carpet & 46.9 & 45.6 & 94.7 & 96.2 & 96.6  & 93.1 & 96 \\
grid & 18.3 & 58.2 & 86.7 & 94.6 & 96  & 87.2 & 89.9 \\
leather & 64.1 & 81.9 & 97.2 & 97.8 & 98.9  & 98.3 & 98.7 \\
tile & 79.7 & 89.7 & 75.6 & 86 & 87.3  & 82.5 & 86 \\
wood & 62.1 & 72.7 & 87.4 & 91.1 & 89.4  & 92.7 & 94.2 \\
bottle & 74.2 & 91 & 95.5 & 94.8 & 96.2  & 95.5 & 96 \\
cable & 55.8 & 82.5 & 90.9 & 88.8 & 92.5  & 90.2 & 89.1 \\
capsule & 30.6 & 86.2 & 93.7 & 93.5 & 95.5  & 94.3 & 96 \\
hazelnut & 84.4 & 91.7 & 95.4 & 92.6 & 93.8  & 94.1 & 96 \\
metal nut & 35.8 & 83 & 94.4 & 85.6 & 91.4  & 92 & 93.8 \\
pill & 46 & 89.3 & 94.6 & 92.7 & 93.2  & 93 & 93.5 \\
screw & 27.7 & 75.4 & 96 & 94.4 & 97.9  & 95.4 & 96.5 \\
toothbrush & 15.1 & 82.2 & 93.5 & 93.1 & 91.5  & 91.6 & 93.7 \\
transistor & 62.8 & 72.8 & 87.4 & 84.5 & 83.7  & 92.9 & 94 \\
zipper & 70.3 & 83.9 & 92.6 & 95.9 & 97.1  & 94.6 & 95.4 \\
\midrule
\textbf{Average}  & \textbf{51.5} & \textbf{79} & \textbf{91.7} & \textbf{92.1} & \textbf{93.4}  & \textbf{92.5} & \textbf{93.9} \\
\bottomrule
    \end{tabular}
    \caption{MVTec Anomaly Segmentation benchmark: PRO}
    \label{tab:pro_benchmarks}
\end{table}

\begin{table}[]
    \centering
    \scriptsize
    \begin{tabular}{ccccccccc}
    \toprule
  & \multicolumn{2}{c}{\textbf{Efficientnet B5}} &  \multicolumn{2}{c}{\textbf{VGG16}} & \multicolumn{2}{c}{\textbf{Resnet18}} & \multicolumn{2}{c}{\textbf{Resnet50}}  \\
  & 1L & 3L & 1L & 3L & 1L & 3L & 1L & 3L \\

 \midrule
carpet & 94.6 & 97.6 & 97.5 & 97.8 & 97.9 & 99.0 & 98.4 & 99.2 \\
grid & 95.0 & 97.3 & 98.2 & 98.5 & 96.7 & 97.4 & 97.3 & 98.1 \\
leather & 98.6 & 99.3 & 98.8 & 99.0 & 99.7 & 99.8 & 99.8 & 99.8 \\
tile & 93.4 & 95.5 & 92.6 & 92.7 & 88.8 & 93.3 & 95.1 & 96.5 \\
wood & 91.7 & 93.8 & 96.2 & 96.3 & 94.6 & 96.6 & 96.7 & 97.7 \\
bottle & 98.6 & 98.3 & 97.9 & 97.6 & 98.4 & 98.6 & 98.7 & 98.8 \\
cable & 96.3 & 96.7 & 95.7 & 95.8 & 95.9 & 96.7 & 97.4 & 97.0 \\
capsule & 98.7 & 98.5 & 99.2 & 99.2 & 98.3 & 98.9 & 98.8 & 99.1 \\
hazelnut & 98.3 & 98.1 & 98.8 & 98.9 & 98.3 & 99.0 & 98.8 & 99.1 \\
metal nut & 96.2 & 95.5 & 96.3 & 96.4 & 95.4 & 97.0 & 96.6 & 97.3 \\
pill & 96.4 & 96.1 & 97.5 & 97.6 & 96.1 & 97.0 & 96.9 & 97.0 \\
screw & 98.4 & 98.1 & 99.5 & 99.6 & 99.1 & 99.3 & 98.9 & 99.2 \\
toothbrush & 98.3 & 97.9 & 98.5 & 98.5 & 98.2 & 98.9 & 98.4 & 98.7 \\
transistor & 95.4 & 96.7 & 95.6 & 95.6 & 95.6 & 96.4 & 96.3 & 96.5 \\
zipper & 96.9 & 97.7 & 98.3 & 98.0 & 98.4 & 98.8 & 98.6 & 98.8 \\
\midrule
\textbf{Average} & \textbf{96.4} & \textbf{97.2} & \textbf{97.4} & \textbf{97.4} & \textbf{96.8} & \textbf{97.8} & \textbf{97.8} & \textbf{98.2} \\
\bottomrule
    \end{tabular}
    \caption{MVTec Anomaly Segmentation: pixel-wise AUROC for FRE across backbones}
    \label{tab:auc_nets}
\end{table}

\paragraph{Layer selection:} 
As already described in Section \ref{sec:Method}, our method can be applied to the features of any set of intermediate layers within a DNN. We do this for a variety of backbone architectures. For concision, the results for Resnet50 are shown in \ref{tab:resnet50_layers} and we include additional results for the remaining in the supplementary material. Across all backbone architectures, we observe that the best anomaly localization performance is attained in layers in the middle of the network -- not too close to either the input or the output.

We further investigate if the anomaly localization performance can be improved by combining the FRE maps from multiple layers. To do this, we generate the FRE maps for the two layers closest to the middle layer of the DNN -- one positioned before and one after the selected middle layer. The combined FRE map, $\mb{M}_{3L}$ (subscript $3L$ indicating 3 layers) is generated by taking a geometric average of the three maps
\begin{equation*}
    \mb{M}_{3L} = \left(\mb{M}_{m-1} \cdot \mb{M}_m \cdot \mb{M}_{m+1} \right)^{1/3}
\end{equation*}
where $m$ is the index of the middle layer. We report both single layer, and three-layer results in Tables \ref{tab:auc_nets} and \ref{tab:pro_nets} for various backbone architectures. We also compare FRE single-layer and three-layer performances against other competitive methods in Tables \ref{tab:auc_benchmarks} and \ref{tab:pro_benchmarks}. We observe that single-layer results far exceed most other methods and are at par with the current state of the art, thereby validating the effectiveness of our approach. However, combining scores using FRE maps from three layers allows us to further improve upon the already impressive single-layer results, enabling us to outperform other methods. One could consider including FRE maps from even more layers; however, these would entail additional computations, and the improvements to AUC and PRO metrics are likely to be increasingly marginal. In real-world implementations, having such flexibility has practical value as it enables the user to choose their configuration based on available compute and desired performance targets.

In Fig. \ref{fig:sample_outputs}, we show the anomaly heatmaps for several sample input images. We see that the FRE maps from the single-layer already correspond very well to the ground truth. A small but noticeable improvement is further observed in the quality of the anomaly heatmaps obtained by fusing FRE maps from three layers, consistent with the improvement in the pixel-level AUC and PRO metrics.

\paragraph{Complexity and Speed:} Next, we present results demonstrating the low computational complexity of our proposed approach. We implemented the full pipeline in our approach on a system comprising only an Intel Xeon Platinum 8280 CPU with no discrete GPU. The results for both model training and run-time during inference for the MVTec dataset, as evaluated on various pretrained DNNs, are presented in Table \ref{tab:complexity}. Here, training time is the time taken to learn the PCA model (not network weights) for features from a single layer on the training split of the anomaly dataset, which in this case is one of the fifteen MVTec categories. For inference, the full pipeline includes image loading, rescaling (to match the required input dimensions required by the backbone DNN), running a forward pass of the DNN to generate the features, and calculating FRE scores from these features. For inference, we report the frame-rate that is achieved by our method when processing input of the same resolution as MVTec images, which are typically $1024\times 1024$ pixels in size. 

As can be seen, our approach is trained in a matter of seconds! Furthermore, our approach achieves very high frame-rates on all DNN backbones. These frame rates exceed the 30fps found on typical industrial cameras. In an industrial setup, therefore, where several cameras might be deployed to detect defects at different stages of the overall manufacturing pipeline, multiple such streams can be serviced by a single CPU unit, making our solution very attractive for practical real-world deployment.

\begin{table}[]
    \centering
    \scriptsize
    \begin{tabular}{ccccccccc}
    \toprule
  & \multicolumn{2}{c}{\textbf{Efficientnet B5}} &  \multicolumn{2}{c}{\textbf{VGG16}} & \multicolumn{2}{c}{\textbf{Resnet18}} & \multicolumn{2}{c}{\textbf{Resnet50}}  \\
  & 1L & 3L & 1L & 3L & 1L & 3L & 1L & 3L \\
 \midrule
carpet & 79.5 & 90.8 & 90.7 & 91.6 & 93.3 & 95.4 & 93.1 & 96.0 \\
grid & 80.8 & 89.1 & 86.6 & 86.5 & 85.1 & 87.5 & 87.2 & 89.9 \\
leather & 93.7 & 96.7 & 94.4 & 95.3 & 98.0 & 98.6 & 98.3 & 98.7 \\
tile & 79.7 & 85.4 & 74.6 & 71.5 & 66.6 & 73.9 & 82.5 & 86.0 \\
wood & 81.6 & 85.7 & 90.5 & 91.2 & 89.6 & 92.6 & 92.7 & 94.2 \\
bottle & 94.4 & 94.5 & 93.4 & 93.6 & 94.6 & 95.8 & 95.5 & 96.0 \\
cable & 87.1 & 89.1 & 83.5 & 84.4 & 87.2 & 88.5 & 90.2 & 89.1 \\
capsule & 93.7 & 93.7 & 95.2 & 96.7 & 93.6 & 95.7 & 94.3 & 96.0 \\
hazelnut & 91.4 & 92.1 & 94.8 & 94.7 & 93.3 & 96.0 & 94.1 & 96.0 \\
metal nut & 89.6 & 89.9 & 91.0 & 90.9 & 90.6 & 93.7 & 92.0 & 93.8 \\
pill & 91.1 & 92.3 & 93.5 & 93.9 & 91.1 & 92.9 & 93.0 & 93.5 \\
screw & 93.7 & 93.8 & 97.8 & 98.1 & 96.0 & 97.2 & 95.4 & 96.5 \\
toothbrush & 91.4 & 90.6 & 93.0 & 93.5 & 92.6 & 94.5 & 91.6 & 93.7 \\
transistor & 92.4 & 95.1 & 91.4 & 91.9 & 91.8 & 92.9 & 92.9 & 94.0 \\
zipper & 90.1 & 92.8 & 93.3 & 91.7 & 93.61 & 95.4 & 94.6 & 95.4 \\
\midrule
\textbf{Average} & \textbf{88.7} & \textbf{91.4} & \textbf{90.9} & \textbf{91.0} & \textbf{90.47} & \textbf{92.7} & \textbf{92.5} & \textbf{93.9} \\
\bottomrule
    \end{tabular}
    \caption{MVTec Anomaly Segmentation: PRO for FRE across backbones}
    \label{tab:pro_nets}
\end{table}

\begin{table}[]
    \centering
    \scriptsize
    \begin{tabular}{cccccc}
    \toprule
 & Layer1 & Layer2 & Layer3 & Layer4 \\ 
 \midrule
\midrule
PRO & \textbf{85.49} & \textbf{92.49} & \textbf{91.33} & \textbf{71.26} \\ 
Pixel AUROC & \textbf{94.69} & \textbf{97.77} & \textbf{97.41} & \textbf{93.20} \\ 
\bottomrule
    \end{tabular}
    \caption{MVTec Anomaly Segmentation: FRE Resnet50 Performance across layers}
    \label{tab:resnet50_layers}
\end{table}

\begin{table}[]
    \centering
    \footnotesize
    \begin{tabular}{c|c|c}
    \toprule
    \midrule
        \multirow{2}{*}{Model} & Inference & Avg. Training \\
         & Frame-rate (fps) & Time (s) \\
    \midrule
    Resnet18  & 251.5 & 5.77 \\
    Resnet50  & 89.2 & 11.25 \\
    EfficientNet-B5  & 61.5 & 7.35 \\
    VGG16  & 133.1 & 8.75 \\
    \bottomrule
    \end{tabular}
    \caption{Inference framerate (frames/sec) and average training time in seconds for each MVTec category on different DNNs as measured on an Intel Xeon Platinum 8280 CPU}
    \label{tab:complexity}
\end{table}

\section{Conclusions and Future Work}
This work presented a fast and principled approach for visual anomaly detection and segmentation. We propose applying linear dimensionality reduction on the intermediate features of a pre-trained DNN, prior to leveraging the \emph{feature reconstruction error} (FRE), the $\ell_2$-norm of the difference between the original feature and the pre-image of its reduced embedding, as an uncertainty score for detection. When applied at convolutional layers, we further derive FRE maps that provide pixel-level spatial localization of the anomalies in the image (i.e. segmentation). Detailed experimentation on popular industrial anomaly detection datasets show qualitative performance at par or better than state-of-the-art methods that are significantly more complex.

{\small
\bibliographystyle{ieee_fullname}
\bibliography{egbib}
}

\end{document}